\documentclass[letterpaper, 10 pt, conference]{ieeeconf}  

\IEEEoverridecommandlockouts                              

\usepackage{graphicx}
\usepackage{amsmath}
\usepackage{amssymb}
\usepackage{newtxtext,newtxmath}

\usepackage{subcaption}
\usepackage{booktabs}
\usepackage{multirow}
\usepackage{array}

\usepackage{url}
\usepackage[symbol]{footmisc}
\usepackage{stfloats}
\usepackage{balance}

\usepackage{pifont}
\usepackage[dvipsnames]{xcolor}
\newcommand{\cmark}{\textcolor{ForestGreen}{\ding{51}}}
\newcommand{\xmark}{\textcolor{BrickRed}{\ding{55}}}

\title{\LARGE \bf SPARC: \underline{S}pine with \underline{P}rismatic \underline{A}nd \underline{R}evolute \underline{C}ompliance\\for Faster Quadrupedal Bounding}

\author{Yue Wang$^{1}$ and Yanran Xu$^{1}$
\thanks{*This work was supported by the School of Electronics and Computer Science, University of Southampton.}
\thanks{$^{1}$Yue Wang and Yanran Xu are with the School of Electronic and Computer Science, University of Southampton, United Kingdom.
        {\tt\small \{yue.wang, y.xu\}@soton.ac.uk}}%
}

\begin{document}

\maketitle
\thispagestyle{empty}
\pagestyle{empty}

\begin{abstract}
Quadruped mammals coordinate sagittal spinal bending with axial extension and compression during dynamic locomotion. Yet most robotic quadrupeds use rigid trunks, passively compliant spines with fixed properties, or actively controlled spines that track prescribed trajectories. Whether actively regulated spinal compliance can support faster dynamic locomotion remains unclear. We present SPARC, a compact 1.26-kg, 3-DoF sagittal-plane spine that combines revolute and prismatic motion with independently tunable task-space stiffness and damping. A floating-base impedance controller renders the desired task-space compliance, and benchtop tests show that the fitted axial stiffness matches commanded values within 1.5\%. We integrate SPARC into an 8-DoF quadruped and evaluate it across 97 bounding trials under three spine configurations: impedance-controlled SPARC, the same SPARC module held near a fixed pose using position control, and a lightweight rigid spine. Impedance-controlled SPARC reaches 1.029~m/s, compared with 0.769~m/s for position-controlled SPARC and 0.673~m/s for the rigid spine. Impedance-controlled SPARC reaches higher speeds with larger axial motion and greater mechanical power exchange, while at matched speed it has a higher electrical cost of transport than the rigid spine, revealing an energetic trade-off. Code and hardware are available at: \url{https://github.com/YueWang996/sparc}
\end{abstract}


\section{INTRODUCTION}
Quadruped mammals combine bending of the spine in the sagittal plane with axial extension and compression during dynamic locomotion \cite{alexander1985elastic,schilling2006sagittal}. These motions can lengthen stride, absorb landing impacts, and generate thrust \cite{alexander1985elastic, zhang2022mechanism}. These kinematics also change with locomotion speed, and flexion contributes to stride length and elastic energy storage \cite{schilling2006sagittal, bertram2009motions}. Most robotic quadrupeds, however, employ rigid trunks to simplify mechanical design and control \cite{kau2019stanford, katz2019mini}, leaving spinal motion unavailable as a controllable degree of freedom during dynamic locomotion.

Existing robotic spines reproduce only parts of these capabilities. Passive compliant spines can store and return elastic energy efficiently, but their mechanical properties are largely fixed by design \cite{zhang2013bio}. Actuated spines under position control can prescribe spinal trajectories, but the commanded motion does not provide independently tunable task-space stiffness and damping \cite{khoramshahi2013benefits}. Spring-loaded prismatic spines can provide axial extension, but are better suited to discrete maneuvers than continuous bidirectional motion over repeated gait cycles \cite{ye2023novel}. Combining spinal bending with axial changes in spine length expands the range of trunk configurations available during bounding, while task-space impedance regulation determines how the spine responds to external forces. Whether a robotic spine that combines both motion modes with real-time, independent regulation of task-space stiffness and damping can support faster bounding remains unclear.

\begin{figure}[t]
    \centering
    \includegraphics[width=\linewidth]{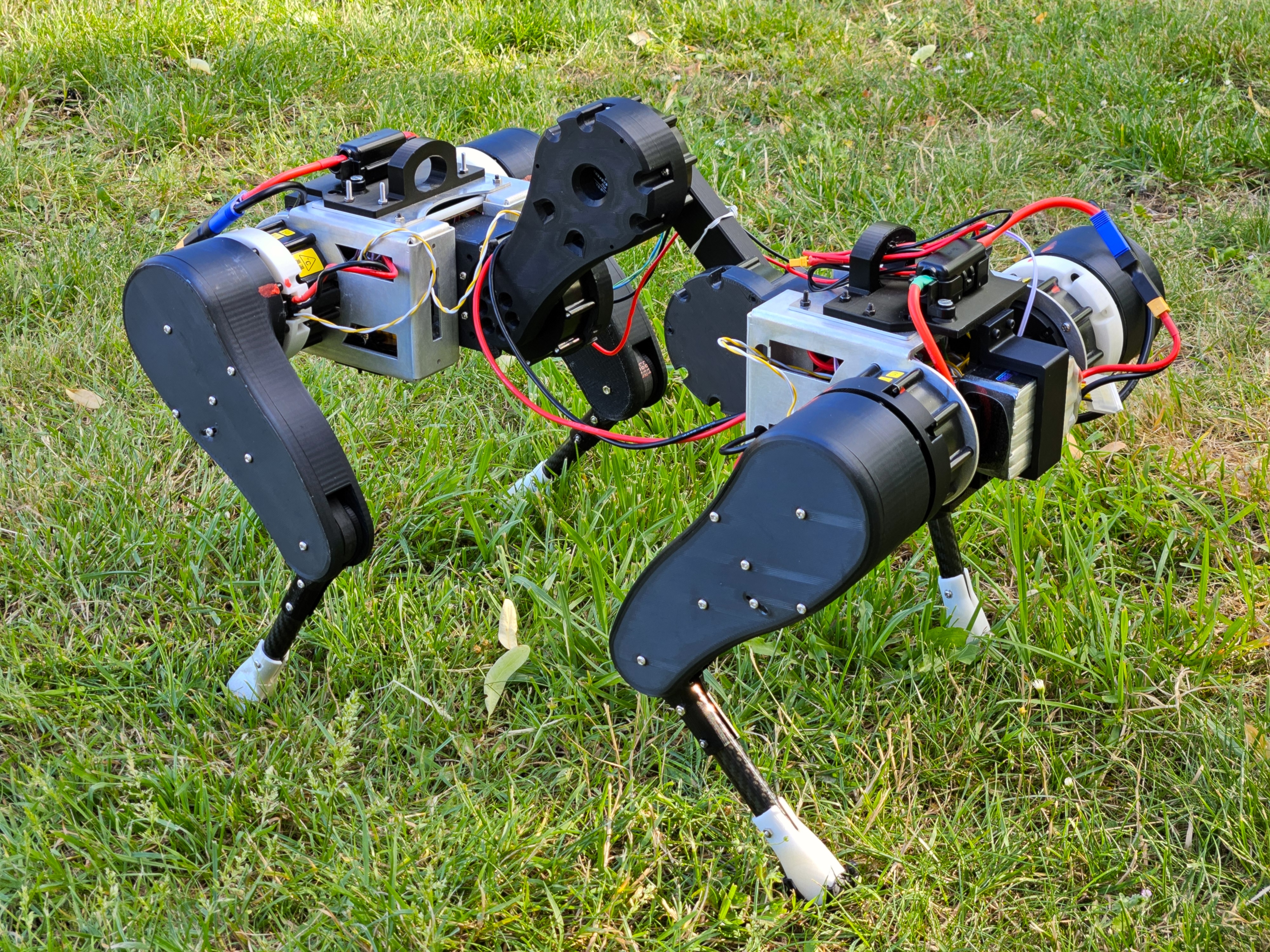}
    \caption{SPARC integrated into the 8-DoF quadruped used for the hardware locomotion experiments. The three-actuator prismatic and revolute spine links the fore and hind body segments and renders task-space compliance.}
    \label{fig:spine_introduction}
    \vspace{-0.6cm}
\end{figure}

To investigate this question, we present \textbf{SPARC} (\textbf{S}pine with \textbf{P}rismatic \textbf{A}nd \textbf{R}evolute \textbf{C}ompliance, Fig.~\ref{fig:spine_introduction}), a compact 3-DoF spine module that combines bending in the sagittal plane with axial extension and compression. SPARC regulates task-space stiffness and damping while allowing the relative pose of the fore and hind bodies to vary during bounding. The main contributions are:

\begin{enumerate}
\item A 1.26~kg self-contained spine in which three quasi-direct-drive (QDD) actuators produce bending in the sagittal plane together with continuous axial extension and compression.

\item A floating-base task-space impedance controller that independently specifies stiffness and damping along the axial, vertical, and pitch axes. The controller evaluates the full floating-base rigid-body dynamics (RNEA/CRBA) at 1~kHz on an embedded microcontroller using a customized lightweight, dependency-free library~\cite{robotdynamicsc}. Benchtop tests show that the fitted axial stiffness matches commanded values within 1.5\%.

\item Full-system hardware experiments across 97 bounding trials under three spine configurations: impedance-controlled SPARC, position-controlled SPARC, and a lightweight rigid spine. Impedance-controlled SPARC reaches the highest measured speed, 1.34$\times$ that of position-controlled SPARC and 1.53$\times$ that of the rigid spine.
\end{enumerate}

The remainder of this paper is organized as follows. Section~\ref{sec:related_work} reviews related work on robotic spines, and Section~\ref{sec:methodology} details the SPARC mechanism and its task-space impedance controller. Section~\ref{sec:results} presents the benchtop characterization and full-system bounding experiments, and Section~\ref{sec:conclusion} concludes.

\begin{table*}[ht]
\centering
\caption{Comparison of spine designs in quadruped robots, grouped by primary spine motion. Online tuning means the impedance can be changed while the robot runs, and independent tuning means separate stiffness and damping along $(x,z,\theta)$.}
\label{tab:compare_spine_features}
\footnotesize
\setlength{\tabcolsep}{5pt}
\begin{tabular}{lccccl}
\toprule
System / Study & \shortstack{Sagittal\\Revolute} & \shortstack{Axial\\Prismatic} & \shortstack{Online\\Tuning} & \shortstack{Independent\\$(x,z,\theta)$} & Key Capability \\
\midrule
\multicolumn{6}{@{}l}{\textit{Sagittal revolute}}\\
Lynx \cite{eckert2015comparing}       & \cmark & \xmark & \xmark & \xmark & Gait modulation \\
Bobcat \cite{khoramshahi2013benefits}   & \cmark & \xmark & \xmark & \xmark & Modest efficiency gains \\
INU Robot \cite{duperret2016core}          & \cmark & \xmark & \xmark & \xmark & Extended workspace \\
Yat-sen Lion \cite{li2023dynamic}             & \cmark & \xmark & \xmark & \xmark & Load distribution \\
Planar Robot \cite{chen2017effect}            & \cmark & \xmark & \cmark & \xmark & Adjustable compliance \\
\addlinespace[2pt]
\multicolumn{6}{@{}l}{\textit{Axial prismatic}}\\
STRAY \cite{wang2024stray}                    & \xmark & \cmark & \xmark & \xmark & Axial twist and extension \\
SCIQ \cite{ye2023novel}  & \xmark & \cmark & \xmark & \xmark & Impact absorption \\
Preloaded Jumping Spine \cite{ye2021modeling} & \xmark & \cmark & --- & --- & Standing long jump \\
\addlinespace[2pt]
\multicolumn{6}{@{}l}{\textit{Axial twist (out of sagittal plane)}}\\
Twisting Spine \cite{caporale2023twisting}      & \xmark & \xmark & \xmark & \xmark & Axial-twist study \\
Solo9 \cite{qian2024robust}                     & \xmark & \xmark & \xmark & \xmark & Steering and robustness \\
\addlinespace[2pt]
\multicolumn{6}{@{}l}{\textit{Sagittal revolute + axial prismatic}}\\
Spine Morphology Study \cite{fisher2017effect} & \cmark & \cmark & --- & --- & Rapid-acceleration morphology \\
\midrule
\textbf{SPARC (this work)}       & \cmark & \cmark & \textbf{\cmark} & \textbf{\cmark} & \textbf{Faster stable bounding} \\
\bottomrule
\end{tabular}
\end{table*}

\section{RELATED WORK}\label{sec:related_work}
The spine of a fast-running mammal such as the cheetah contains many vertebrae rather than a single joint, and their articulations are loosely constrained, especially in the lumbar region, so each contributes part of the total deflection \cite{randau2017morphological}. The resulting motion is therefore not solely rotational: a galloping cheetah flexes and extends its spine in the sagittal plane while its fore and hind bodies also approach and separate along the body axis \cite{alexander1985elastic,schilling2006sagittal,bertram2008motions}. A robot spine that reproduces this needs both sagittal revolute and axial prismatic motions. It also requires a controller that regulates the stiffness and damping of these motions, since the back stores and returns elastic energy across a stride \cite{alexander1985elastic}.

Sagittal rotation is the motion robotic spines usually provide. It has been used to extend leg workspace \cite{duperret2016core}, to improve efficiency \cite{khoramshahi2013benefits, chen2017effect, bhattacharya2019learning, culha2012actuated}, and to coordinate the fore and hind bodies \cite{li2023dynamic}. None of these designs lengthen or shorten along the body axis, which can be advantageous for rapid acceleration. Fisher \emph{et al.} optimized planar trajectories for robots drawn at random from a design space and compared rigid, revolute and prismatic morphologies. In their simulations, a prismatic spine was the most effective at rapid acceleration for 75\% of the robots, against 18\% for rigid and 6\% for revolute \cite{fisher2017effect}.

Prismatic spines have also been built. Ye and Karydis co-optimized the spring parameters and takeoff trajectory of a preloaded and lockable elastic prismatic spine for a standing long jump in simulation \cite{ye2021modeling}, and their SCIQ platform later realized such a spine in hardware \cite{ye2023novel}. In both, the spring is preloaded manually before the run, and the locking mechanism releases it at a single instant during the motion, so the axial extension is one discrete energy release rather than a continuously regulated behavior. STRAY instead provides a spine that extends and retracts continuously, driven by a reinforcement-learning policy \cite{wang2024stray}. However, it couples that motion with axial twisting rather than with the sagittal rotation that accompanies axial length change in the mammalian spine. Axial twisting has itself been studied as a spine motion in its own right, for steering and disturbance rejection \cite{caporale2023twisting,qian2024robust}, though it acts out of the sagittal plane. In neither case, though, is the spine's stiffness or damping something the controller sets: the spring fixes them in hardware, and the learned policy leaves them implicit in a trajectory.

Adjustable spine compliance has been built before. Passive and mechanically variable-stiffness spines return elastic energy well \cite{zhang2013bio,eckert2015comparing,pouya2017spinal}, but the stiffness is fixed at assembly or adjusted slowly by a dedicated mechanism, and the damping is whatever the structure provides. One sagittal spine modulates its compliance online \cite{chen2017effect}, and quasi-direct-drive (QDD) actuators put joint impedance under software control \cite{di2018dynamic}. These approaches, however, define compliance through the particular mechanism or actuator used. A revolute joint, an elastic element, and a prismatic actuator require different control variables and parameters, so there is no common joint-level description of spine compliance. All of them, however, change the pose of the hind body relative to the fore body. SPARC therefore specifies stiffness and damping directly on that relative planar pose, $(x,z,\theta)$, and coordinates its three joints to realize the commanded impedance. Task-space impedance control is well established in manipulation, but it has rarely been applied to a robotic spine.

Table~\ref{tab:compare_spine_features} compares SPARC with representative spine designs on four properties: sagittal rotation, axial length change, impedance that can be tuned while the robot runs, and stiffness and damping set independently along $(x,z,\theta)$. To our knowledge, SPARC is the first robotic spine to demonstrate all four in hardware.

\section{SPARC DESIGN AND CONTROL}\label{sec:methodology}
\begin{figure}[h]
    \centering
        \includegraphics[width=0.9\linewidth]{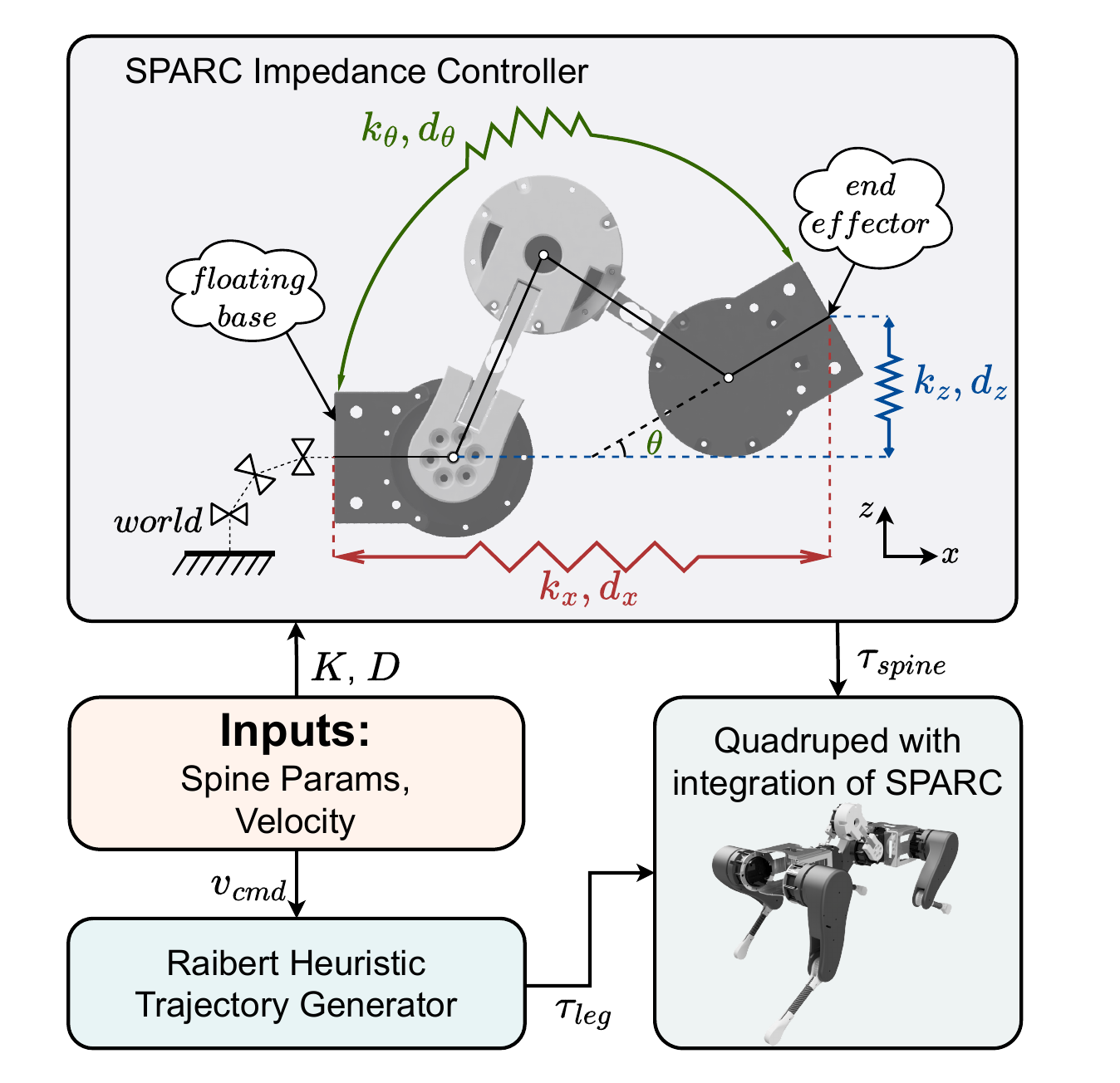}
    \caption{System integration of SPARC. The impedance controller regulates task-space compliance ($K, D$), while a Raibert-based heuristic~\cite{raibert1986legged} generates bounding trajectories for the quadruped testbed.}
    \label{fig:control-arch}
\end{figure}

SPARC addresses these gaps with two components: a 3-DoF serial spine that provides revolute and prismatic motion, and a task-space impedance controller that independently regulates stiffness and damping. The spine module is integrated between the fore and hind bodies of an 8-DoF quadruped, while a bounding gait generator produces the leg trajectories for the robot, as shown in Fig.~\ref{fig:control-arch}. The following subsections describe the spine hardware, impedance controller, and real-time embedded implementation.

\subsection{Hardware Design}
SPARC is a self-contained module comprising a 3-DoF active spine, a Motion Control Board (MCB), and a Power Management Unit (PMU). Three Xiaomi CyberGear QDD actuators drive the sagittal-plane motion, each with a peak torque of 12~N$\cdot$m. Their axes are parallel and normal to the sagittal plane, so the spine is a planar serial chain, and rotating the three joints together translates the end-effector along the axial direction without a sliding element. The controller holds a 0.22~m nominal axial length and clamps the commanded pose to a 0.18--0.27~m range. The links are 3D-printed and light, and 75\% of the 1.26~kg module mass sits near the joint axes, which keeps the end-effector inertia low, so the QDD actuators remain transparent to interaction forces. Table~\ref{tab:spine_params} lists the link masses, segment lengths, and inertias used by the dynamics model. 

The middle joint has the shortest moment arm about the axial line of action of the end-effector and therefore sets the axial force limit. This moment arm is 88~mm at the nominal pose, sso its peak torque gives up to 137~N along the axial direction, which corresponds to a rendered axial stiffness of up to about 2900~N/m over the commanded range.

A custom MCB built around an STM32G473 microcontroller runs the low-level control loops. The three spine actuators share a single CAN bus channel. The board carries a 6-axis IMU (ICM-42688-P) for attitude estimation and SPI and I2C interfaces to an upstream processor. The impedance loop closes on the actuator encoders. A custom PMU adds electronic-fuse protection with current and voltage sensing and fast disconnect of the DC bus.

\begin{table}[t]
\centering
\caption{Physical parameters of the spine links used by the dynamics model.}
\label{tab:spine_params}
\footnotesize
\setlength{\tabcolsep}{4pt}
\begin{tabular}{lccc}
\toprule
Link & \shortstack{Mass\\(kg)} & \shortstack{Segment\\(mm)} & \shortstack{$I_{xx},I_{yy},I_{zz}$\\(kg$\cdot$mm$^2$)} \\
\midrule
Hind body   & 0.377 & 4.5   & 183, 333, 234 \\
Hind spine  & 0.399 & 117.5 & 195, 536, 476 \\
Front spine & 0.052 & 117.5 & 10, 84, 82 \\
Front body  & 0.377 & 58.7  & 183, 333, 234 \\
\bottomrule
\end{tabular}
\end{table}

Load cells are mounted in the central links and measure the rendered end-effector force directly instead of inferring it from motor current. We use this force reference in the benchtop validation of Section~\ref{sec:static-validation} and remove the load cells for the locomotion trials. The module links the fore and hind body segments of a custom 8-DoF quadruped whose legs use a parallel joint arrangement with belt-driven knees to keep leg inertia low. The leg motors track the joint angles produced by the bounding-gait scheduler through a gain-scheduled PD controller~\cite{di2018dynamic}, and the complete robot has a mass of 6.66~kg.

\subsection{Impedance Control of SPARC}
We developed a model-based impedance controller to command the spine as a compliant, spring-like mechanism. SPARC is modeled as a floating-base system in the sagittal plane, with the base link fixed to the hind body segment and the spine end-effector to the fore body.

We model the floating-base system with the equation of motion~\cite{featherstone2008rigid}
\begin{equation}
M(q)\dot{v} + h(q,v) = \tau + J^\top F_\text{ext},
\end{equation}
where $M(q)\in\mathbb{R}^{9\times9}$ is the mass matrix, $h(q,v)=C(q,v)+g(q)$ collects the Coriolis and gravity effects, $\tau\in\mathbb{R}^9$ are the generalized torques, and $F_\text{ext}$ are external forces with Jacobian $J$. The configuration and velocity are $q = [q_{\text{b}}^\top\ q_{\text{s}}^\top]^\top \in \mathbb{R}^{10}$ and $v = [v_{\text{b}}^\top\ v_{\text{s}}^\top]^\top \in \mathbb{R}^{9}$, where $q_{\text{b}} \in \mathbb{R}^{7}$ stacks the base position and its unit quaternion, $v_{\text{b}} \in \mathbb{R}^{6}$ is the base linear and angular velocity, and $q_{\text{s}}, v_{\text{s}} \in \mathbb{R}^{3}$ are the spine joint positions and velocities. CRBA and RNEA compute $M$ and $h$ at run time. The IMU measures the base orientation and the actuator encoders measure $q_{\text{s}}$ and $v_{\text{s}}$.

We divide the mass matrix into blocks corresponding to the base and spine coordinates,
\begin{equation}
M(q) = \begin{bmatrix} M_{bb}(q) & M_{bs}(q) \\ M_{sb}(q) & M_{ss}(q) \end{bmatrix},
\end{equation}
where $M_{bb}\in\mathbb{R}^{6\times6}$ is the base inertia, $M_{ss}\in\mathbb{R}^{3\times3}$ is the spine inertia, and $M_{sb}=M_{bs}^\top\in\mathbb{R}^{3\times6}$ captures the inertial coupling between base and spine. We likewise partition the nonlinear vector $h=[h_b^\top,\,h_s^\top]^\top$, the generalized torques $\tau=[\mathbf{0}_6^\top,\,\tau_s^\top]^\top$, and the Jacobian $J=[J_b,\,J_s]$. Extracting the spine rows gives
\begin{equation}\label{eq:spine_dyn}
M_{sb}(q)\dot{v}_{\text{b}} + M_{ss}(q)\dot{v}_{\text{s}} + h_{\text{s}}(q,v) = \tau_{\text{s}} + J^\top_{\text{s}} F_\text{ext}.
\end{equation}

The controller does not compensate for the coupling term $M_{sb}(q)\dot v_{\text{b}}$ in \eqref{eq:spine_dyn}. Compensating for it enforces error dynamics of the form $\ddot{e}=-k_pe-k_d\dot{e}$, which cancels the spine's inertial lag during impacts and makes the module a feedback-linearizing tracker of its commanded pose. Leaving it uncompensated lets base acceleration act on the spine's own inertia and produce the lag of a physical compliant element. The term stays on the right-hand side as an input to the spine,
\begin{equation}\label{eq:spine_plant}
M_{ss}(q)\dot{v}_{\text{s}} + h_{\text{s}}(q,v) = \tau_{\text{s}} + J^\top_{\text{s}} F_\text{ext} - M_{sb}(q)\dot{v}_{\text{b}}.
\end{equation}
The term $h_{\text{s}}$ compensates for gravity and Coriolis effects; without it, the rendered stiffness becomes configuration-dependent and the equilibrium drifts. Because the IMU measures the base orientation, gravity is recomputed in the base frame at every cycle, so the rendered impedance stays aligned with the robot's body axes at any base attitude.

As shown in Fig.~\ref{fig:control-arch}, the controlled task-space pose is $\xi = [x, z, \theta]^\top \in \mathbb{R}^3$, the axial (horizontal in the base frame) position, vertical position, and pitch angle of the end-effector relative to the base. It follows from $q_{\text{s}}$ by forward kinematics, with $\dot{\xi} = J_{\text{s}} v_{\text{s}}$. The desired compliant behavior imposes a virtual mass--spring--damper system in task space,
\begin{equation}\label{eq:desired_system}
    \Lambda(\xi)\ddot{\tilde{\xi}} + D\dot{\tilde{\xi}} + K\tilde{\xi} = -F_{\text{ext}},
\end{equation}
where $\tilde{\xi} = \xi_d - \xi$ is the task-space tracking error and $\Lambda(\xi) = (J_{\text{s}} M_{\text{ss}}^{-1} J^\top_{\text{s}})^{-1}$ is the operational-space inertia. We use diagonal stiffness and damping matrices,
\begin{equation}
K = \text{diag}(k_x, k_z, k_\theta), \quad D = \text{diag}(d_x, d_z, d_\theta),
\end{equation}
so that impedance is regulated independently along the axial, vertical, and pitch axes.

The impedance control law follows as
\begin{equation}
\label{eq:impedance}
    \begin{aligned}
        F_{\text{imp}} &= K\tilde{\xi} + D\dot{\tilde{\xi}}, \\
        F_{\text{comp}} &= \Lambda(\xi) \dot{J}_{\text{s}}(q,v)v_{\text{s}}, \\
        \tau_{\text{s}} &= h_{\text{s}}(q,v) + J^\top_{\text{s}}(q)(F_{\text{imp}} - F_{\text{comp}}).
    \end{aligned}
\end{equation}
Here $F_{\text{imp}}$ renders the restorative stiffness and damping and $F_{\text{comp}}$ cancels the task-space kinematic bias. Substituting \eqref{eq:impedance} into the spine dynamics~\eqref{eq:spine_plant} and projecting into task space gives
\begin{equation}\label{eq:closed_loop}
    \Lambda\ddot{\tilde{\xi}} + D\dot{\tilde{\xi}} + K\tilde{\xi} = -F_{\text{ext}} + \Lambda J_{\text{s}} M_{\text{ss}}^{-1} M_{sb}\dot{v}_{\text{b}}.
\end{equation}
The last term is the base excitation that the design retains. On the fixed testbed of Section~\ref{sec:static-validation} the base does not accelerate, and \eqref{eq:closed_loop} reduces to the target model~\eqref{eq:desired_system}. During bounding the term drives the spine as an external input instead of being canceled by the actuators. Damped least squares regularizes $\Lambda$ near singular configurations~\cite{nakamura1986inverse}. The controller preserves the spine's intrinsic operational-space inertia rather than shaping it to a diagonal target, which would require the high-fidelity endpoint force sensing of full inertia shaping~\cite{albu2003cartesian}.

\subsection{Friction Compensation}
Since joint friction limits torque fidelity on real hardware, we compensate for these effects using a smooth Stribeck model~\cite{na2018adaptive} applied element-wise to each joint:
\begin{equation}\label{eq:friction}
\begin{aligned}
\tau_{f,i} &= \Big[\tau_{c,i}+(\tau_{stat,i}-\tau_{c,i})
e^{-(|v_{s,i}|/v_{str})^{\delta}}\Big] \\
&\quad \tanh(\beta v_{s,i}) + b_i v_{s,i},
\end{aligned}
\end{equation}
where $\tau_{c,i}$, $\tau_{stat,i}$, and $b_i$ denote the Coulomb friction, static friction, and viscous coefficient for the $i$-th joint, respectively. The parameter $v_{str}$ is the Stribeck velocity threshold, while $\delta$ and $\beta$ are shape parameters governing the transition smoothness. This formulation ensures a smooth transition between static and Coulomb friction, minimizing chattering near zero velocity while accounting for the nonlinearities observed in QDD actuators. Finally, the total control input sent to the spine actuators is the sum of the impedance and friction compensation torques:
\begin{equation}
\tau_{\text{cmd}} = \tau_{\text{s}} + \tau_{\text{f}}.
\end{equation}

\subsection{Software and Real-Time Implementation}
The controller runs as a high-priority FreeRTOS task at 1~kHz on the STM32G473 microcontroller of the MCB, a 170~MHz Cortex-M4 with an FPU. We developed a customized library to solve the rigid-body dynamics, since established libraries such as Pinocchio~\cite{carpentier2019pinocchio} and RBDL~\cite{felis2017rbdl} are heavy and target desktops rather than MCUs. It uses single-precision arithmetic, static memory allocation, and CMSIS-DSP, and follows the standard spatial-algebra formulation~\cite{featherstone2008rigid} with matrix-multiply-free joint transforms~\cite{wang2026batched}. Each cycle updates the kinematics and $J_{\text{s}}$, evaluates $M$ and $h$ with CRBA and RNEA, and computes \eqref{eq:impedance} and \eqref{eq:friction} in approximately 290~$\mu$s, leaving a 70\% margin within the 1~ms period.
\section{EXPERIMENTS AND ANALYSIS}\label{sec:results}
%
\begin{figure}[t]
\centering
\includegraphics[width=\linewidth]{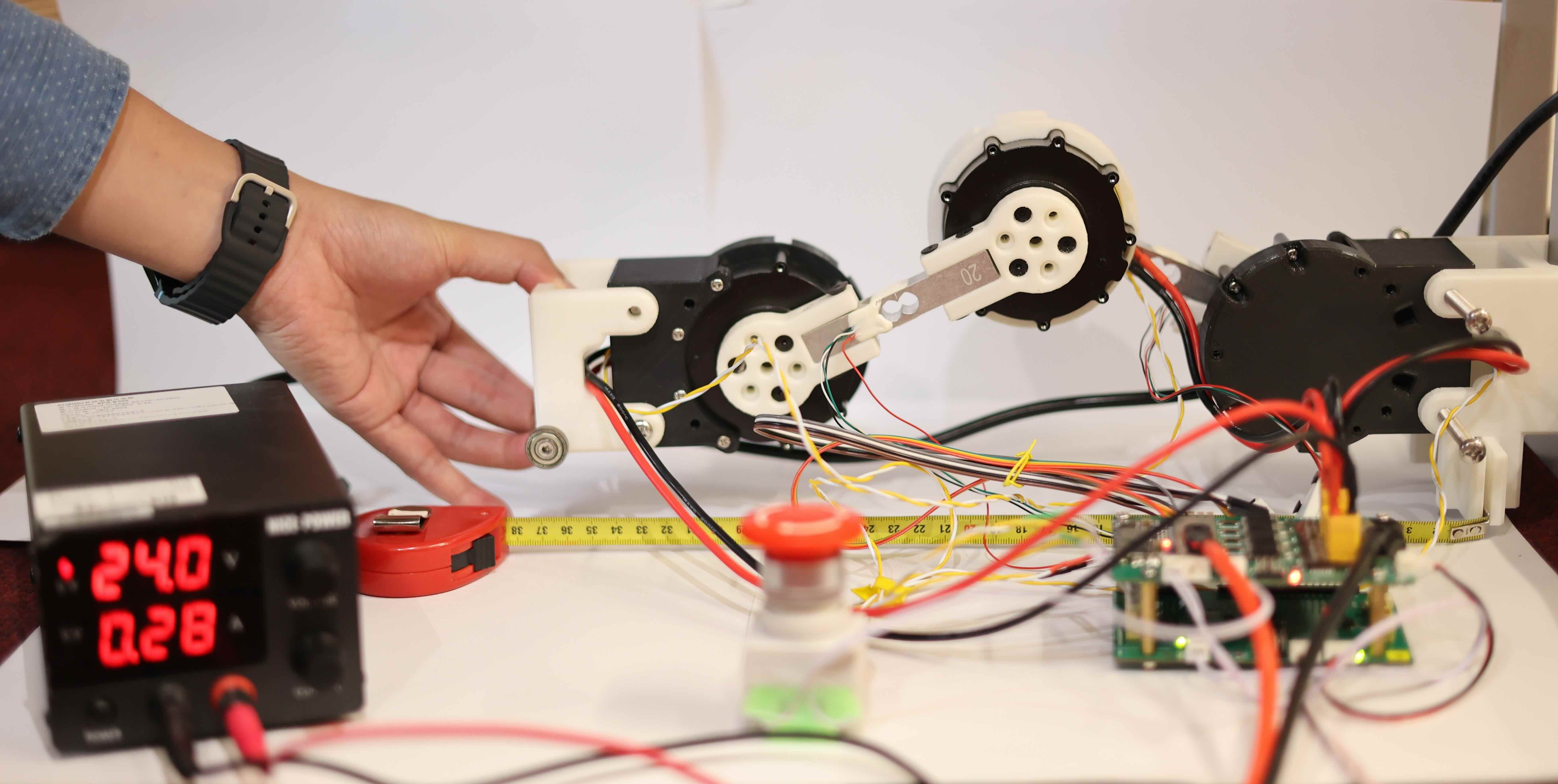}
\caption{Testbed for the SPARC spine unit. The base is fixed, while the end-effector is mechanically supported yet unconstrained along the $x$-axis to minimize friction. An axial force $F_x$ is applied manually for quasi-static tests. The ruler provides a visual reference, while the precise displacement used for analysis is computed via encoder-based forward kinematics.}
\label{fig:testbed}
\end{figure}
\subsection{Validation of Stiffness Rendering}\label{sec:static-validation}
We evaluated the accuracy of the commanded axial stiffness $k_x$ through quasi-static push-pull tests. We manually displaced SPARC's end-effector along the $x$-axis while a load cell measured the restorative force $F_x$, and computed the corresponding displacement from the joint encoders by forward kinematics (Fig.~\ref{fig:testbed}). For each commanded stiffness $k_x \in \{300, 400, 500, 600, 700\}~\mathrm{N/m}$ we ran 10 continuous push-pull cycles. The tested stiffness range is limited by the force a human operator can apply repeatably over the $\pm60$~mm stroke.

\begin{figure}[h]
  \centering
  \begin{subfigure}[t]{0.33\linewidth}\includegraphics[width=\linewidth]{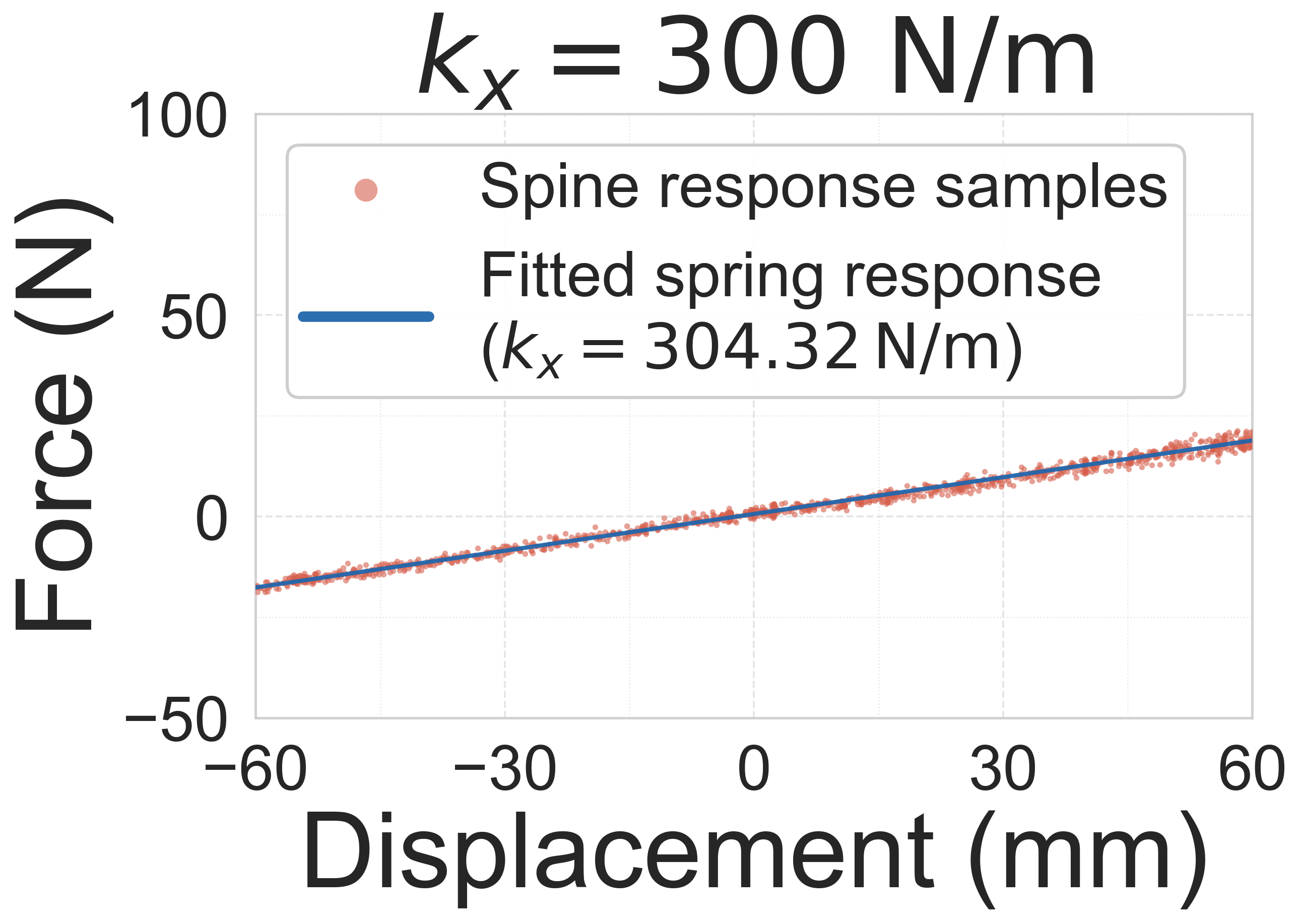}\label{fig:static-k300}\end{subfigure}\hfill
  \begin{subfigure}[t]{0.33\linewidth}\includegraphics[width=\linewidth]{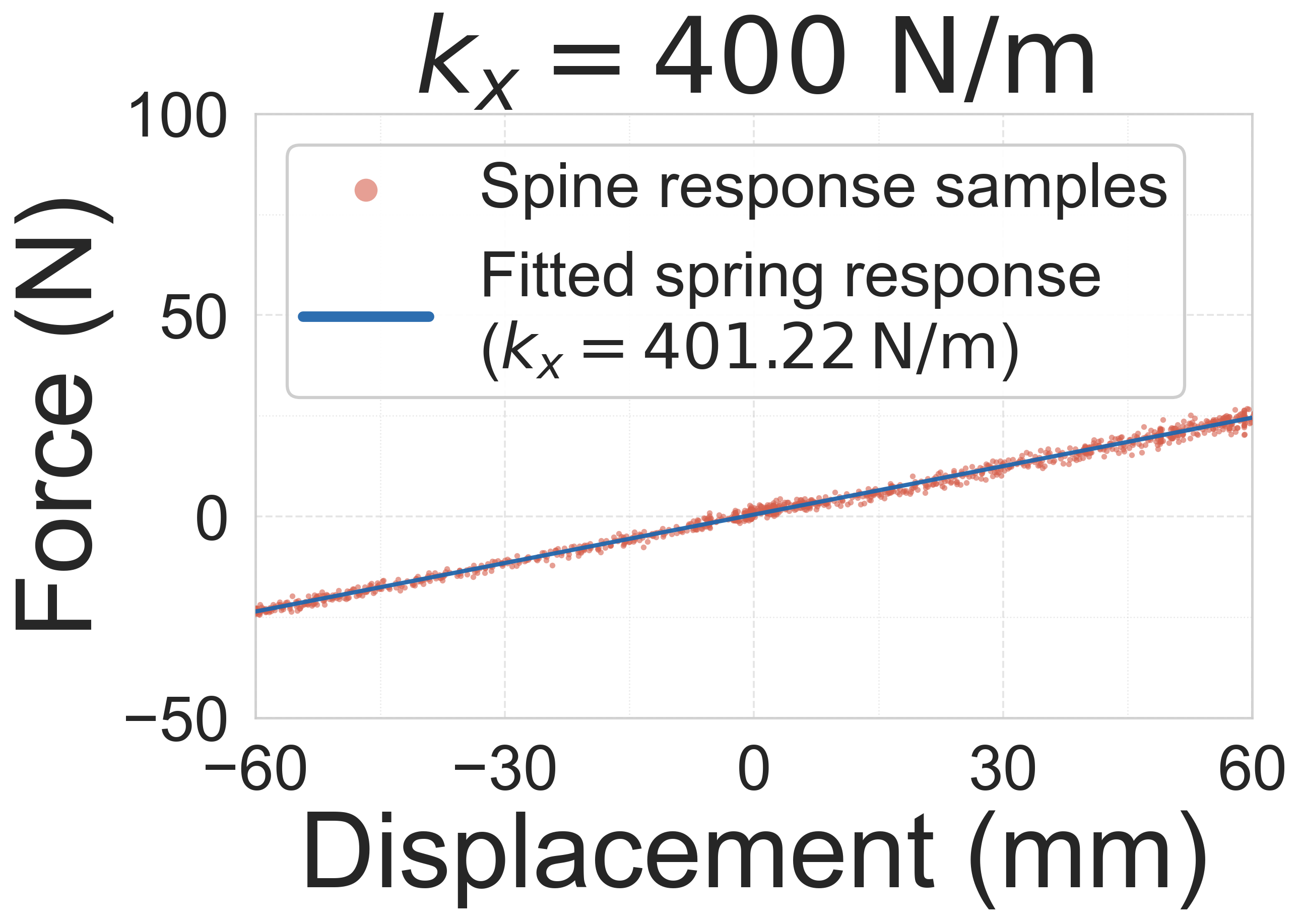}\label{fig:static-k400}\end{subfigure}\hfill
  \begin{subfigure}[t]{0.33\linewidth}\includegraphics[width=\linewidth]{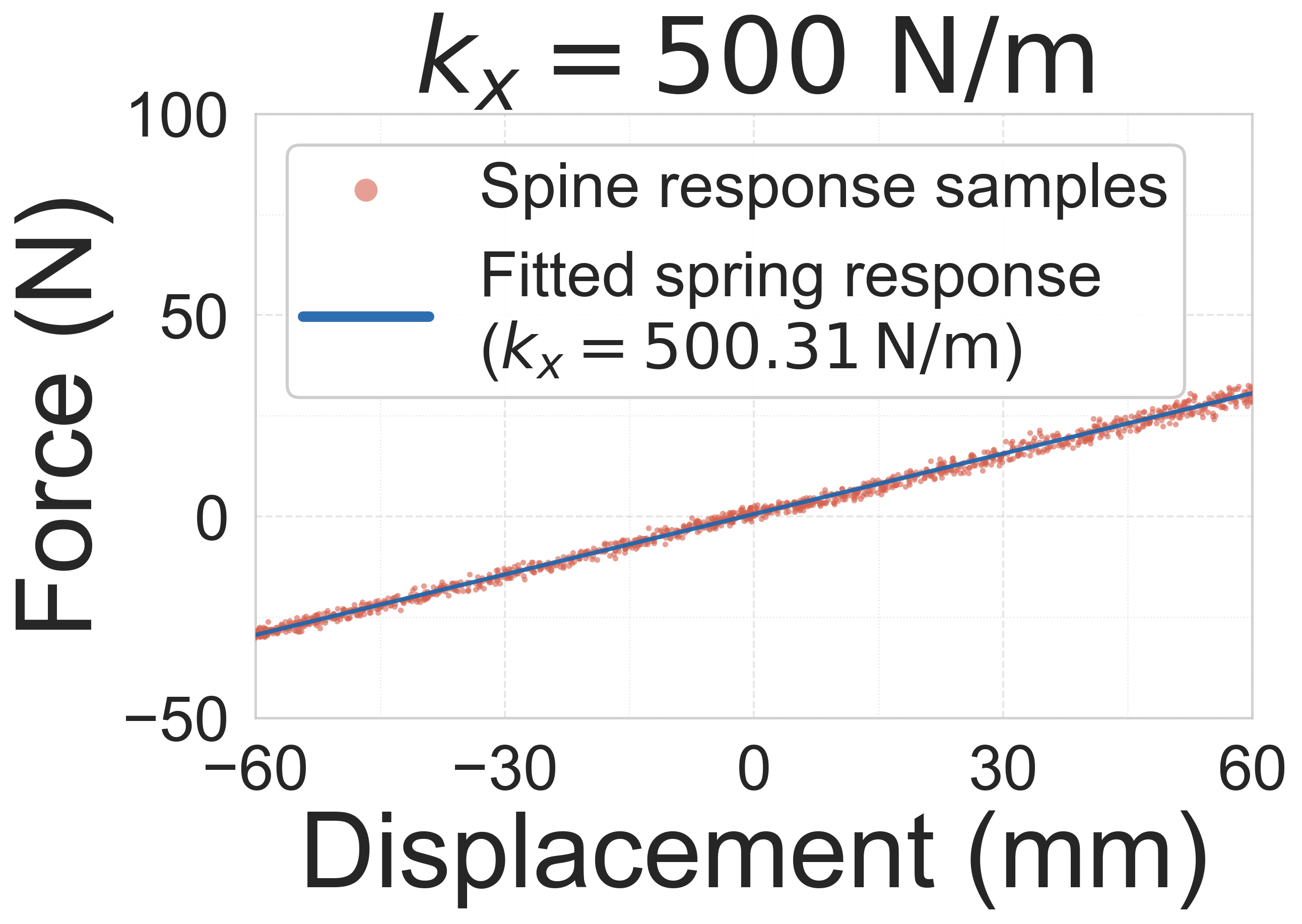}\label{fig:static-k500}\end{subfigure}
  \vspace{0.2cm}
  \begin{subfigure}[t]{0.32\linewidth}\includegraphics[width=\linewidth]{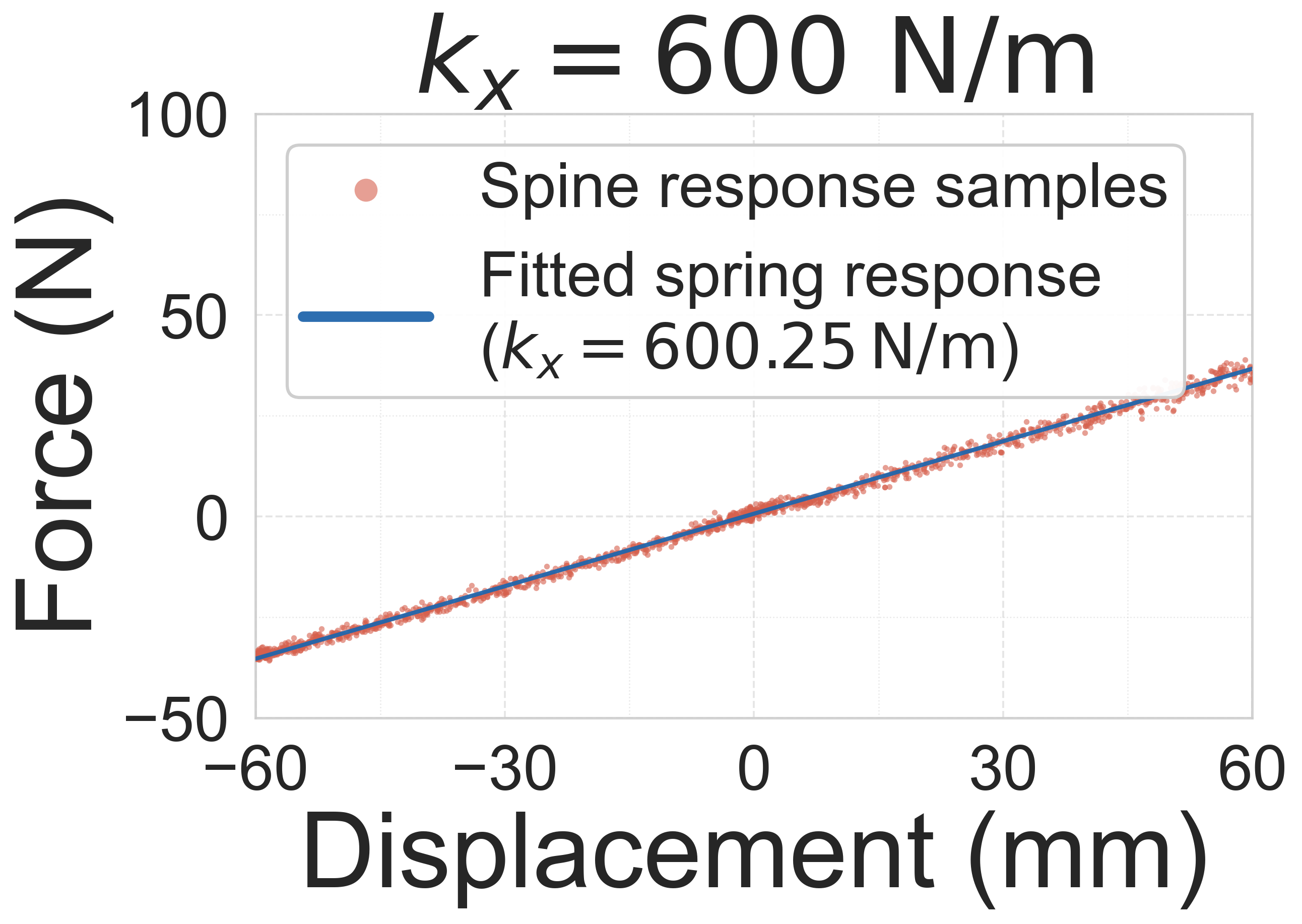}\label{fig:static-k600}\end{subfigure}\hspace{0.05\linewidth}
  \begin{subfigure}[t]{0.32\linewidth}\includegraphics[width=\linewidth]{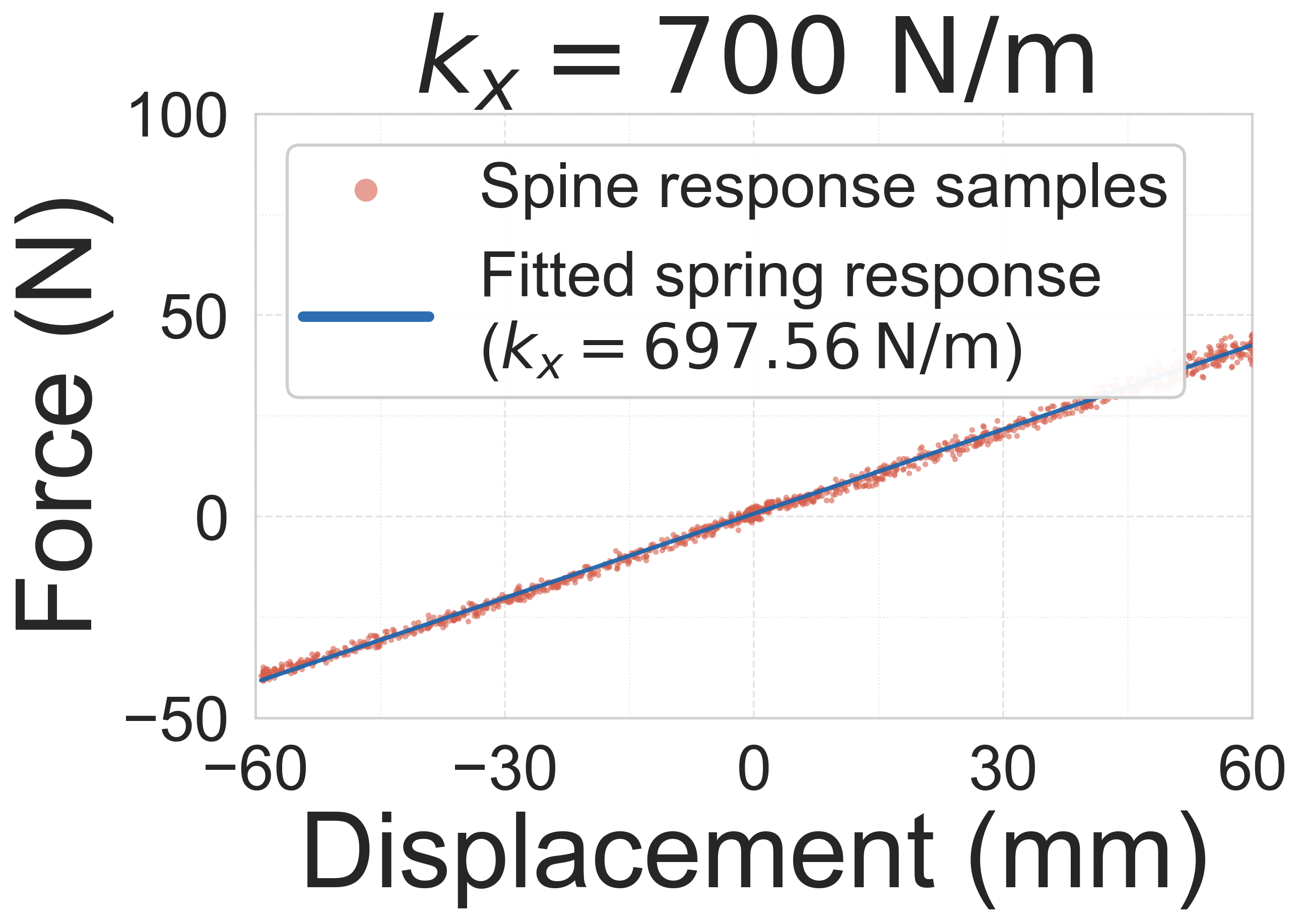}\label{fig:static-k700}\end{subfigure}
  \caption{Static force-displacement response for commanded $k_x\in\{300,400,500,600,700\}\,\mathrm{N/m}$. Dots show measured data and solid lines show linear fits.}
  \label{fig:stiffness-validation}
\end{figure}

Figure~\ref{fig:stiffness-validation} shows the force-displacement responses across all tested stiffness settings. The measured data are highly linear, with a coefficient of determination $R^2 \geq 0.992$ in every case. The fitted stiffness values match the commanded targets with a relative error below 1.5\%, which demonstrates high fidelity in stiffness rendering. We also commanded the vertical and pitch axes at $k_z \in \{300, 600, 900\}$~N/m and $k_\theta \in \{3, 6, 9\}$~N$\cdot$m/rad, and the supplementary video shows the corresponding responses. As a baseline, a model-free PD controller $\tau_{\text{s}}=J^\top_{\text{s}}(K\tilde{\xi}+D\dot{\tilde{\xi}})$ can qualitatively modulate compliance, but under the same tests it exhibits up to 39\% stiffness tracking error due to uncompensated gravity and configuration-dependent kinematic bias.

\subsection{Dynamical Damping Behavior}\label{sec:damping-validation}
For free-response tests, we displaced the end-effector 37~mm and released it. We tested $k_x\in\{300,500\}$~N/m and $d_x\in\{0,2,20,40\}$~N$\cdot$s/m, with 10 repetitions per setting. Because the task-space inertia $\Lambda(\xi)$ varies with configuration, we fitted one constant effective mass jointly across all eight settings while holding stiffness and damping at their commanded values, which provides a linear second-order reference.

\begin{figure*}[thpb]
  \centering
  \begin{subfigure}[t]{0.24\textwidth}
    \includegraphics[width=\linewidth]{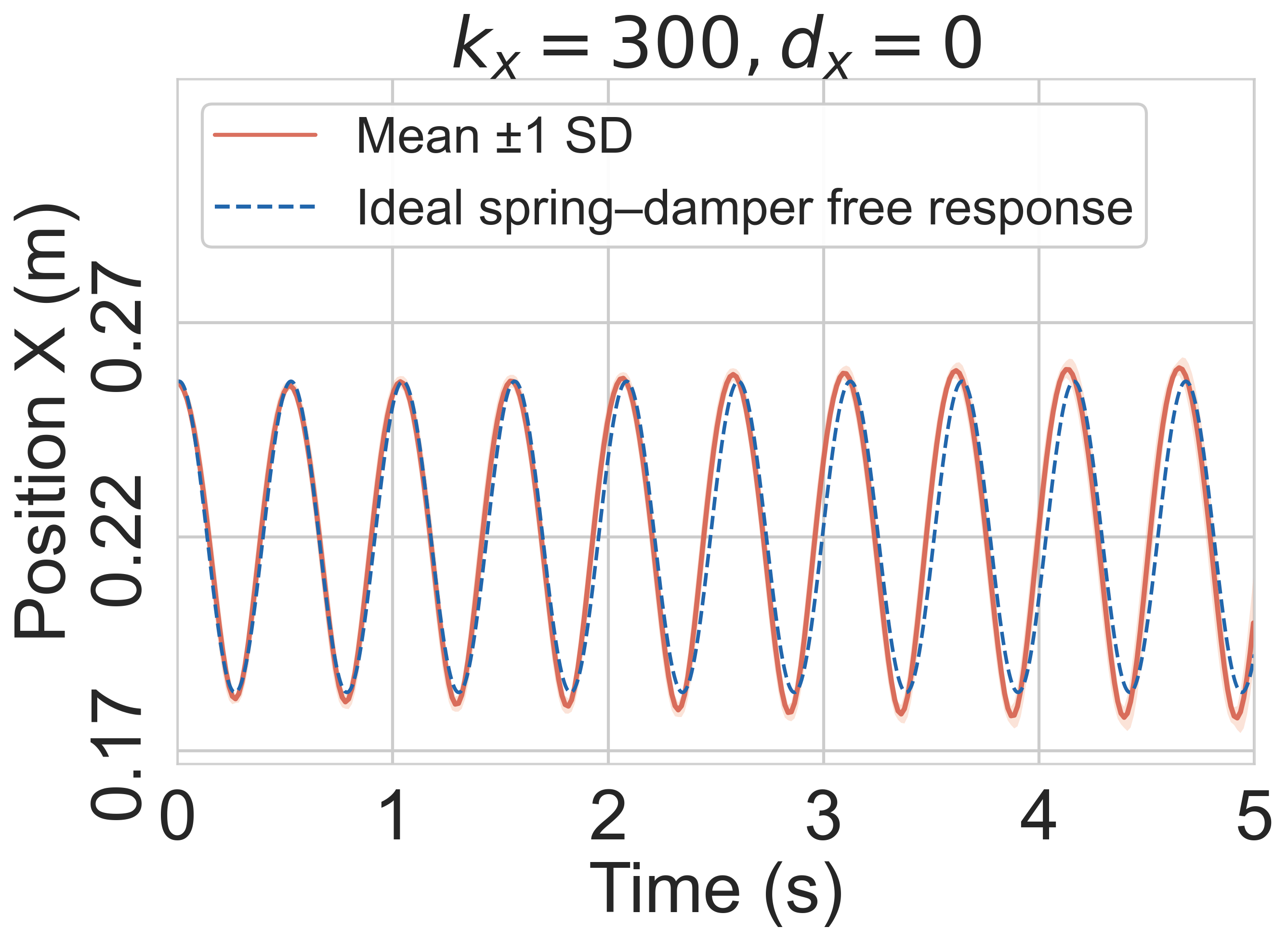}
    \label{fig:k300b0}
  \end{subfigure}\hfill
  \begin{subfigure}[t]{0.24\textwidth}
    \includegraphics[width=\linewidth]{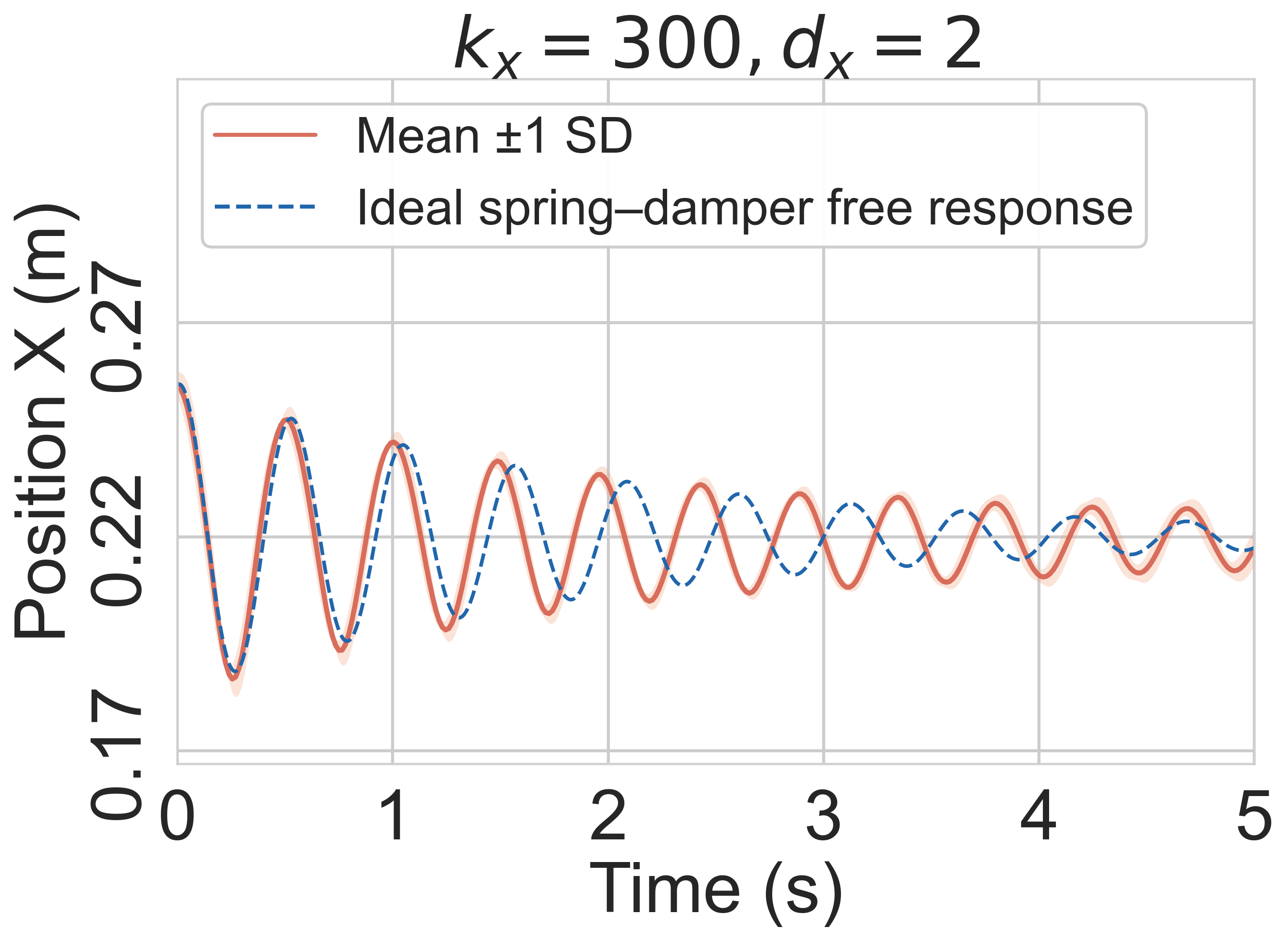}
    \label{fig:k300b2}
  \end{subfigure}\hfill
  \begin{subfigure}[t]{0.24\textwidth}
    \includegraphics[width=\linewidth]{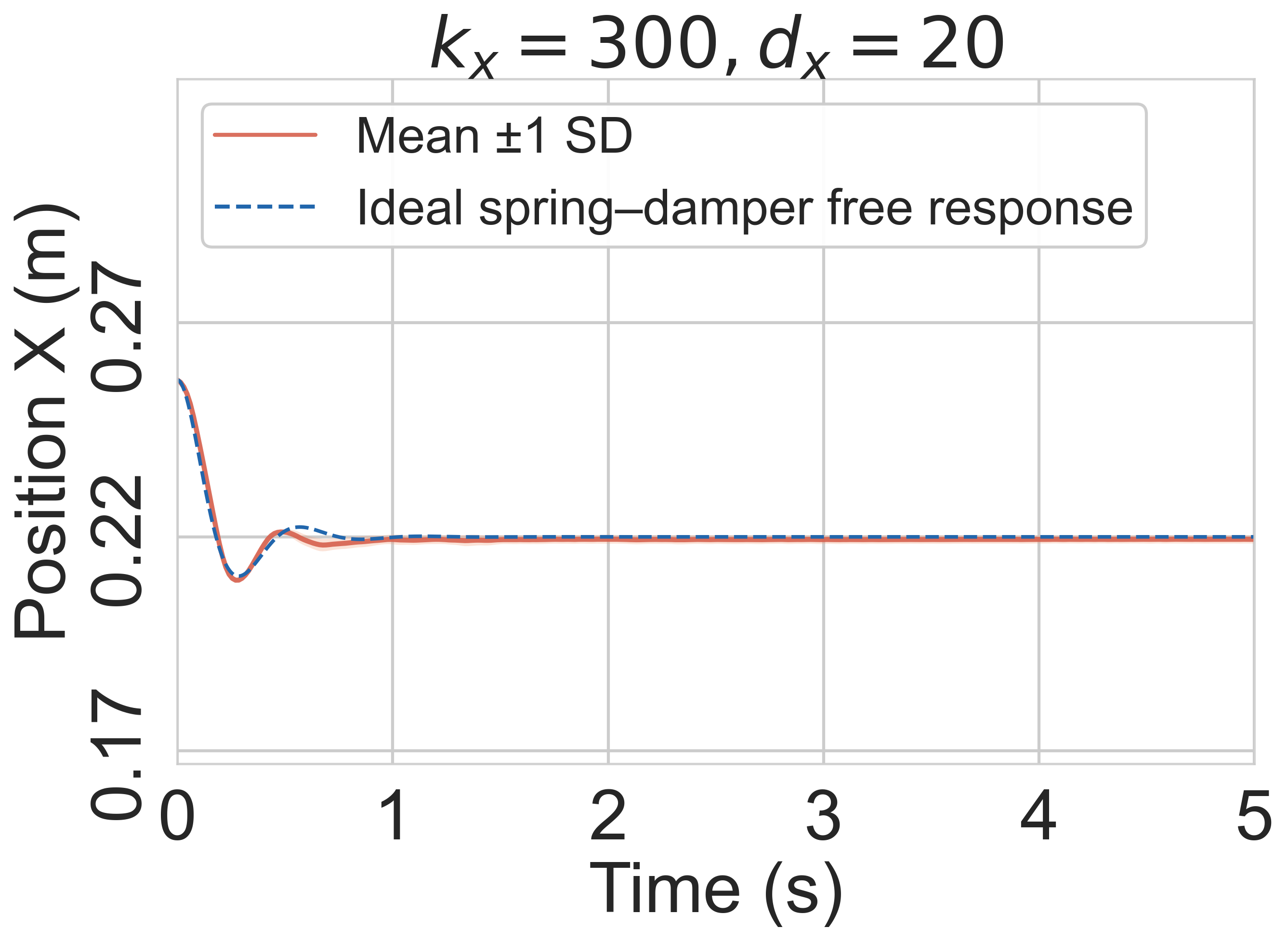}
    \label{fig:k300b20}
  \end{subfigure}\hfill
  \begin{subfigure}[t]{0.24\textwidth}
    \includegraphics[width=\linewidth]{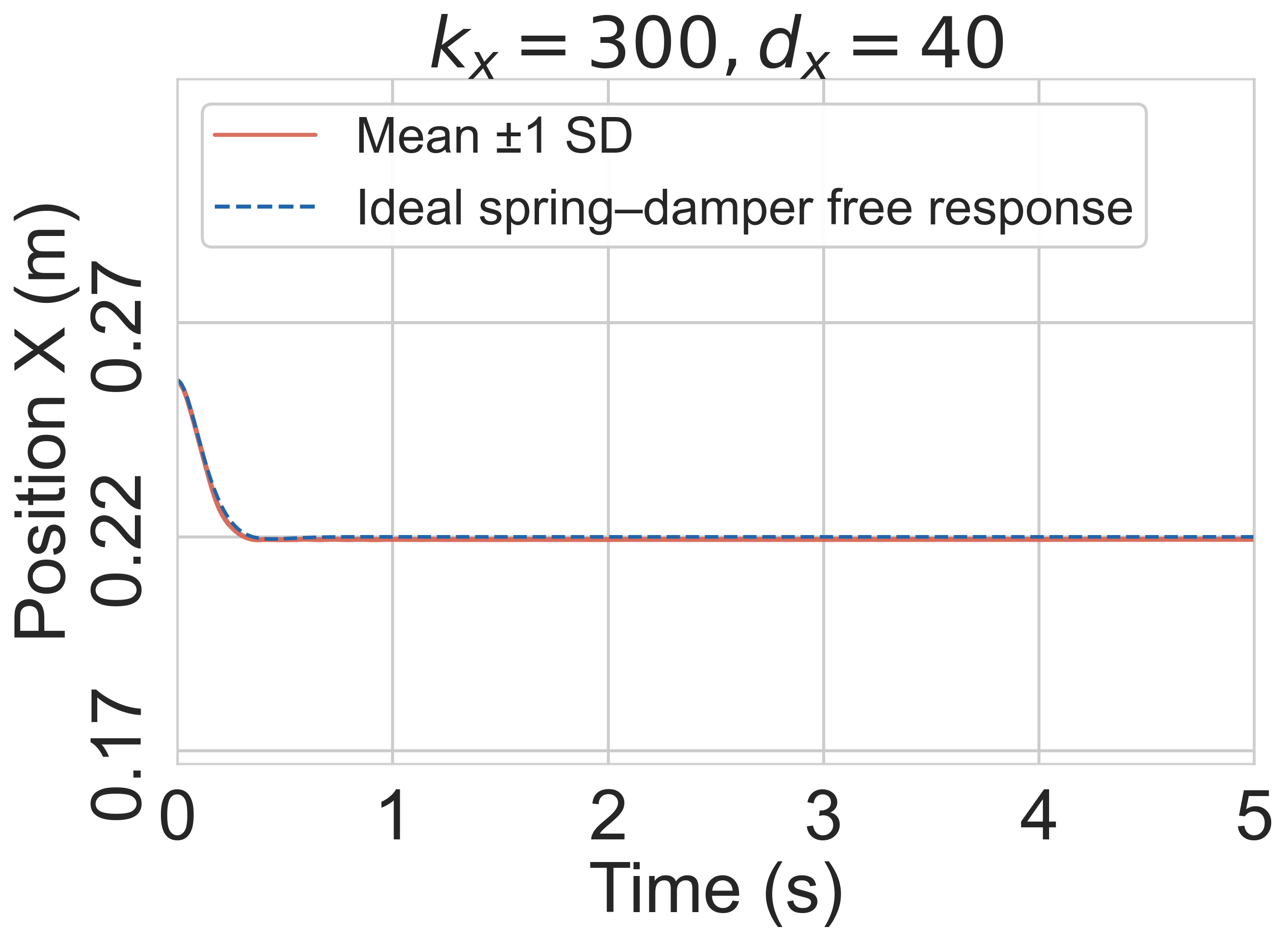}
    \label{fig:k300b40}
  \end{subfigure}

  \begin{subfigure}[t]{0.23\textwidth}
    \includegraphics[width=\linewidth]{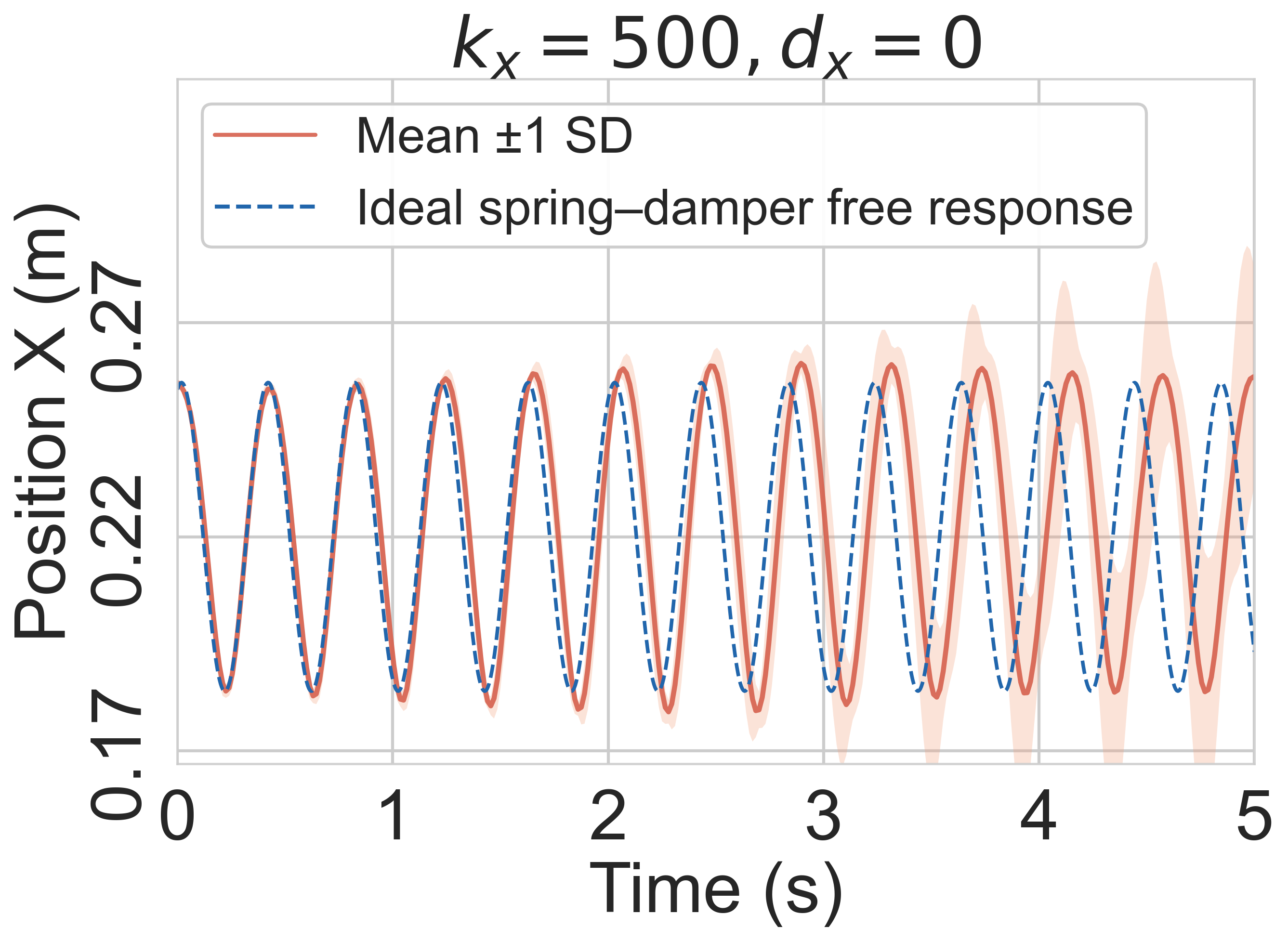}
    \label{fig:k500b0}
  \end{subfigure}\hfill
  \begin{subfigure}[t]{0.23\textwidth}
    \includegraphics[width=\linewidth]{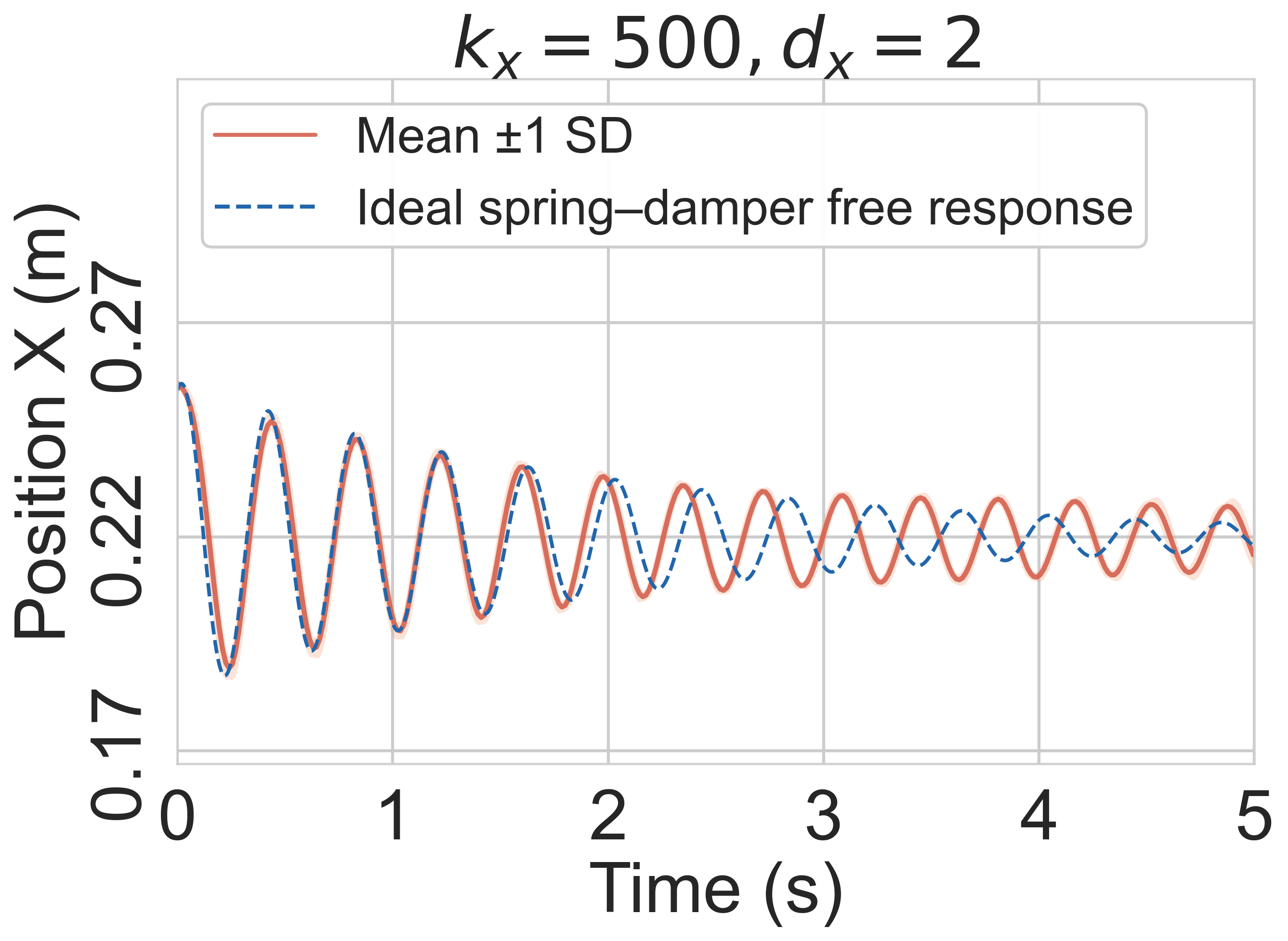}
    \label{fig:k500b2}
  \end{subfigure}\hfill
  \begin{subfigure}[t]{0.23\textwidth}
    \includegraphics[width=\linewidth]{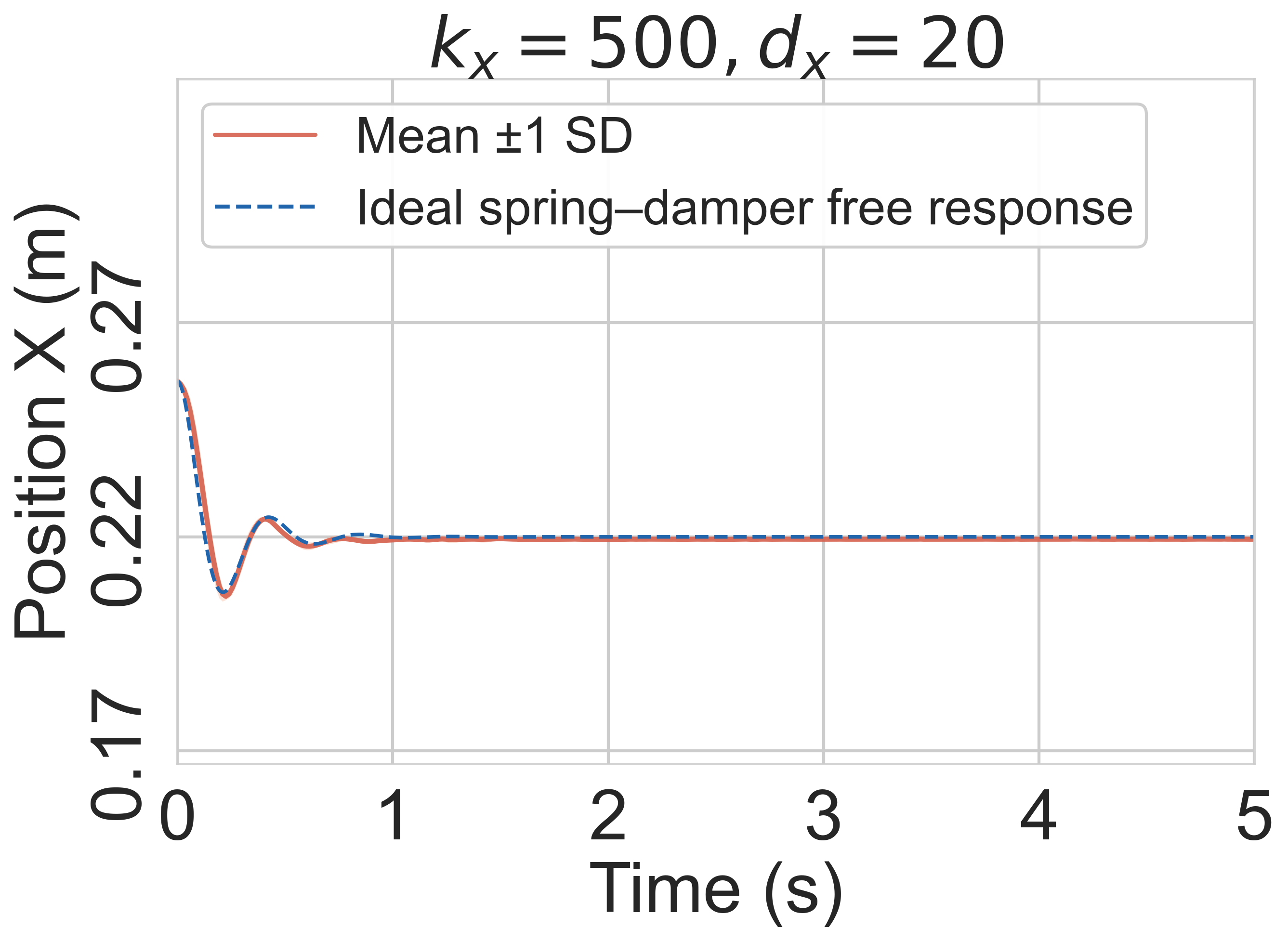}
    \label{fig:k500b20}
  \end{subfigure}\hfill
  \begin{subfigure}[t]{0.23\textwidth}
    \includegraphics[width=\linewidth]{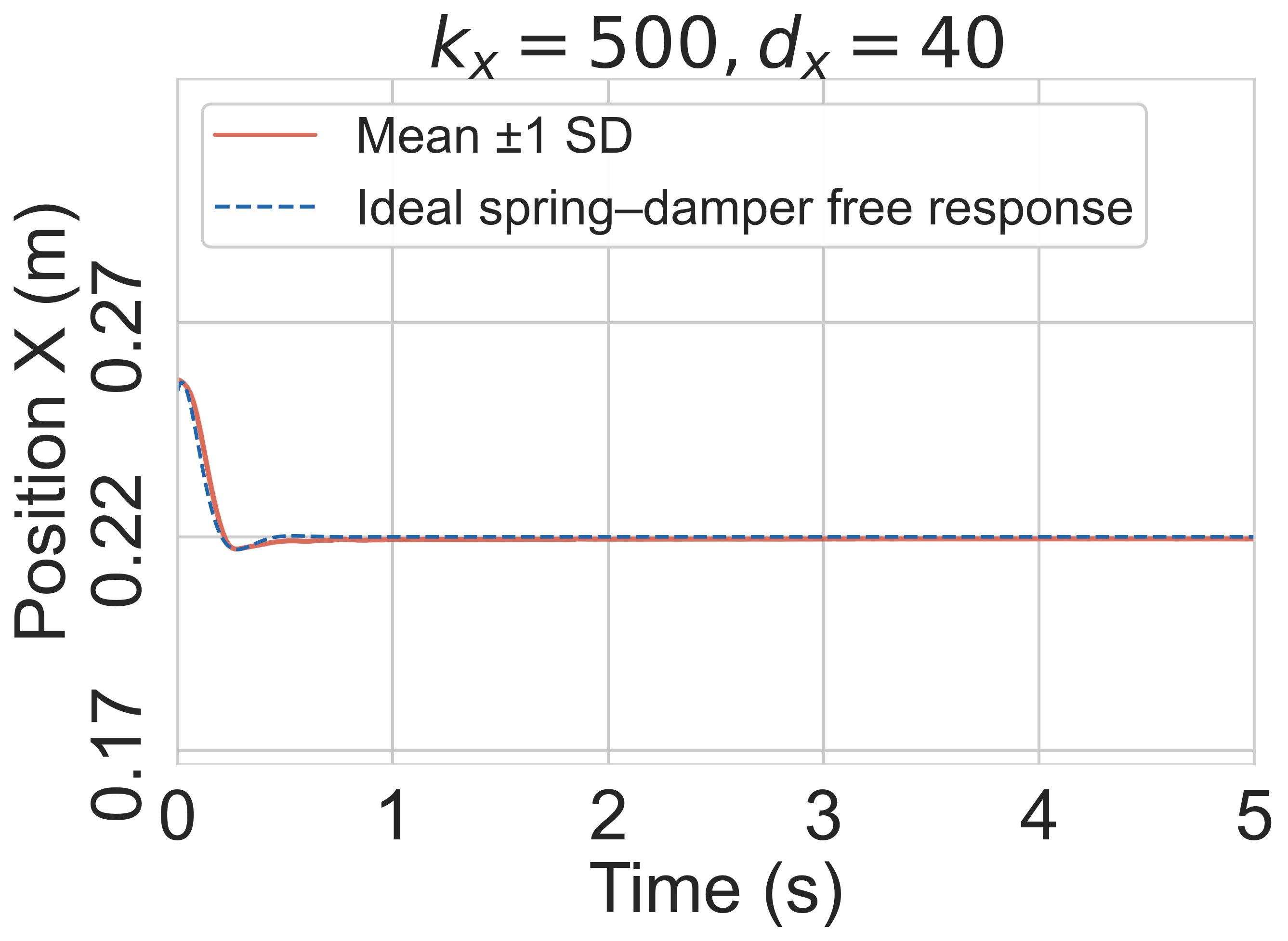}
    \label{fig:k500b40}
  \end{subfigure}
  \vspace{-0.3cm}
  \caption{Experimental validation of the variable impedance controller. The plots display the step response for two stiffness values ($k_x \in \{300, 500\}~\mathrm{N/m}$) under four distinct damping coefficients ($d_x \in \{0, 2, 20, 40\}~\mathrm{N{\cdot}s/m}$). The measured response (solid line, mean $\pm1\sigma$ over 10 trials) is compared against a linear mass-spring-damper reference (dashed line) that uses one effective mass fitted jointly across all eight settings.}
  \label{fig:damping-validation}
\end{figure*}

Figure~\ref{fig:damping-validation} shows that $d_x=20$ and 40~N$\cdot$s/m closely track the fitted reference. The undamped response reveals phase and amplitude drift due to configuration-dependent inertia and friction-model error, while $d_x=2$~N$\cdot$s/m remains underdamped. Across both stiffness settings, increasing commanded damping consistently reduces oscillation and settling time. Together with the static error below 1.5\%, these tests show that the controller systematically and observably modulates the rendered impedance.

\subsection{Locomotion Tests with the SPARC-Integrated Quadruped}\label{sec:protocol}
We integrated the SPARC module between the fore and hind bodies of an 8-DoF quadruped executing a bounding gait and evaluated three spine configurations. For \emph{impedance-controlled SPARC}, the complete 6.66~kg robot carries the three spine actuators and tracks a neutral task-space pose through \eqref{eq:impedance}, and we varied the impedance parameters across runs. For \emph{position-controlled SPARC}, we set the spine motors to position mode to hold the joints stationary, which gives a direct comparison with impedance-controlled SPARC on the same mechanism and the same mass. The electrical power consumed to hold this position is excluded from the efficiency analysis. For the \emph{rigid spine}, we replaced SPARC with a 276~g 3D-printed part, which lowers the robot mass to 5.68~kg and represents a conventional quadruped. The three configurations differ in their feasible gait dynamics, so gait parameters were tuned separately for each condition to obtain stable bounding.

Each run was recorded from gait initiation until the robot stopped, with the electrical power logged at 200~Hz at the same time. The bounding speed was the crossing distance divided by the crossing duration, and the power data were trimmed to the same interval for analysis. Logged power summed the motor-side $dq$ estimates $1.5(V_q i_q + V_d i_d)$, separated into leg and spine-drive components where those channels were available, and was used to compute the electrical cost of transport (CoT)\footnote{$\mathrm{CoT}_{\mathrm{elec}} = \bar{P}_{\mathrm{elec}}/(mgv)$, where $m$ is the total robot mass in each configuration (6.66~kg with SPARC, 5.68~kg with the rigid spine) and $v$ is the measured speed.}.

We analyze 97 bounding trials with steady locomotion across the three spine configurations: 69 with impedance-controlled SPARC, 10 with position-controlled SPARC, and 18 with the rigid spine. Every run has a valid total power measurement, and the separated leg and spine channels are available for 53 impedance-controlled and 8 position-controlled runs. Within the impedance-controlled configuration, a targeted sweep covers 23 distinct $(k_x, d_x)$ settings spanning 1600--2400~N/m and 9--30~N$\cdot$s/m at a fixed gait frequency, and the remaining runs sample a broader range of impedance and gait settings.

\begin{figure}[t]
    \centering
    \includegraphics[width=\linewidth]{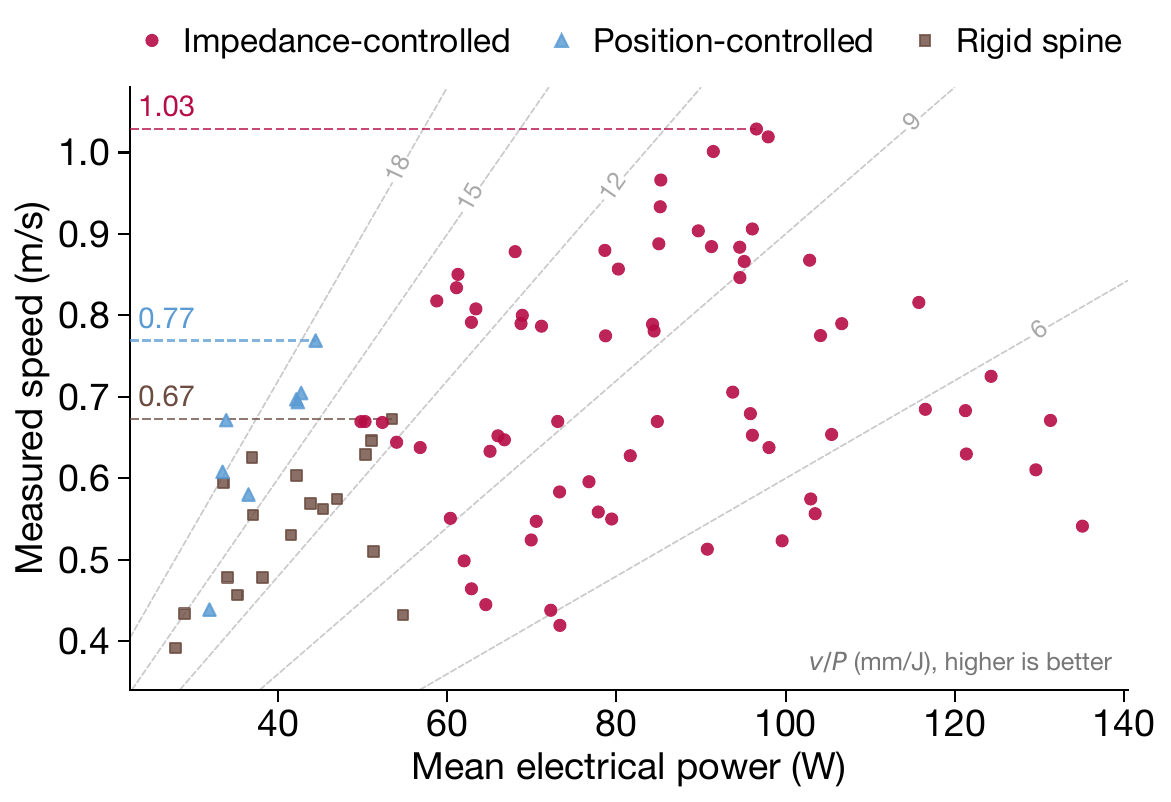}
    \caption{Mean electrical power versus measured speed, one marker per hardware run. For impedance-controlled SPARC, this includes both the leg and the spine motors. Position-controlled SPARC carries the same mass but counts leg power alone, since its spine motors only hold position and would be replaced by rigid material in practice. The lighter rigid spine draws power through its leg motors only. Colored dashed lines mark the fastest run of each condition, and gray contours are lines of constant distance per unit electrical energy ($v/P$, mm/J).}
    \label{fig:power-speed-envelope}
\end{figure}

Figure~\ref{fig:power-speed-envelope} summarizes mean electrical power and measured speed across the three spine configurations. Impedance-controlled SPARC reaches a maximum speed of 1.029~m/s, \textbf{1.34$\times$} the maximum of 0.769~m/s for position-controlled SPARC and \textbf{1.53$\times$} the maximum of 0.673~m/s for the rigid spine. The mean speeds are 0.713, 0.627, and 0.541~m/s, respectively. Of the 69 impedance-controlled SPARC runs, 30 exceed the maximum speed of position-controlled SPARC and 35 exceed that of the rigid spine.

The fast runs are also the ones in which the spine moves most. Between the 10 slowest and the 10 fastest runs, the axial excursion grows from 40 to 123~mm, the axial velocity from 0.42 to 0.83~m/s, and the mechanical power exchanged by the spine from 12.2 to 24.1~W. This speed comes at an energetic cost. At a matched speed of 0.6~m/s, the electrical CoT predicted by condition-wise regression is 2.15 for impedance-controlled SPARC versus 1.34 for the rigid spine.

\subsection{Mechanism of Spine--Leg Power Exchange}\label{sec:cycle}
\begin{figure}[t]
    \centering
    \includegraphics[width=\linewidth]{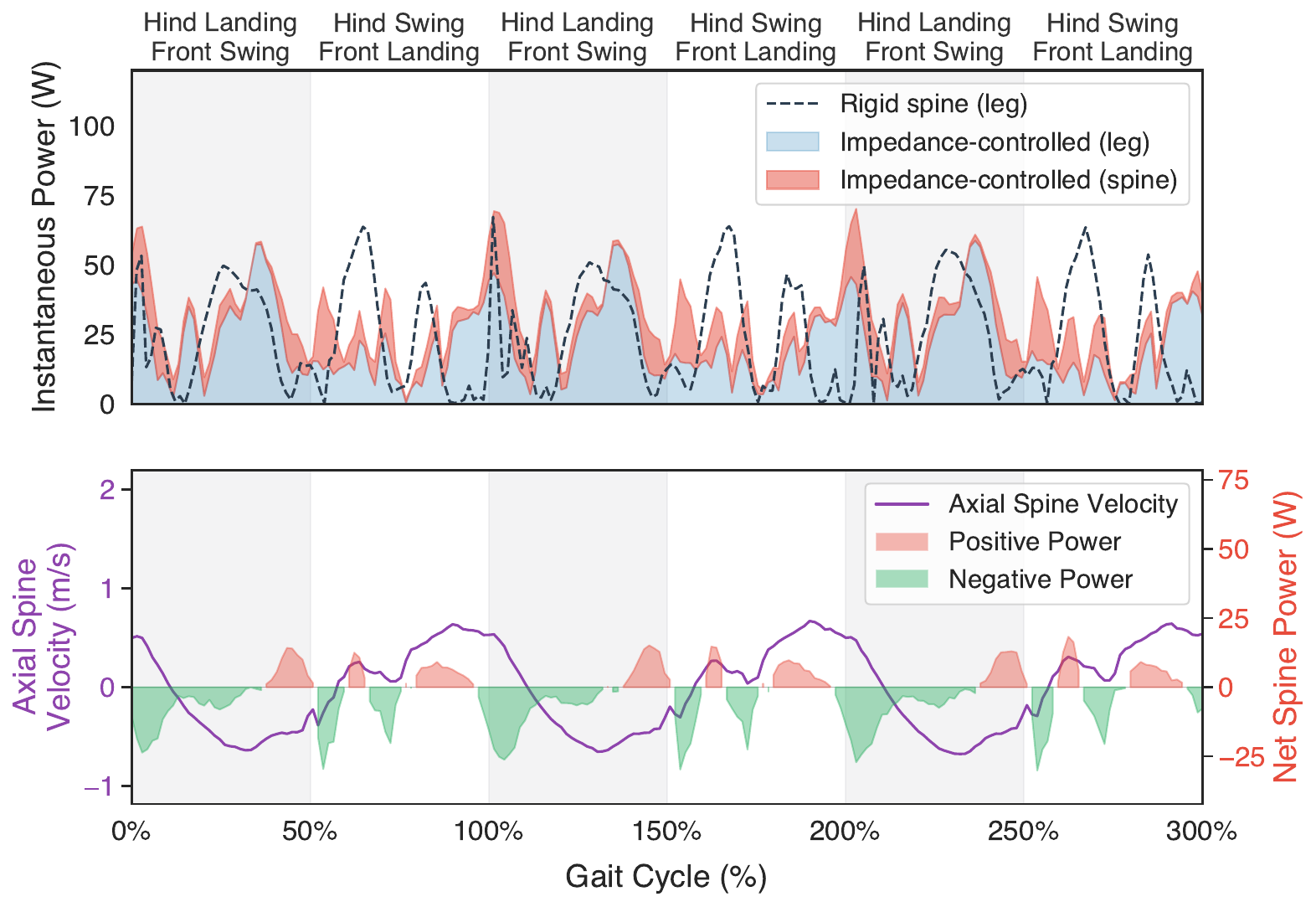}
    \caption{Phase-resolved comparison of impedance-controlled SPARC (0.818~m/s, 58.75~W) and the rigid spine (0.673~m/s, 53.46~W). Three cycles are normalized by gait phase and circularly shifted to align them, without rescaling amplitude. The gray bands mark nominal half-cycles. The upper subplot shows the magnitude of the aggregate leg and spine mechanical power, with the rigid-spine leg trace dashed. The lower subplot shows the axial velocity of impedance-controlled SPARC and its signed power, where positive power is supplied and negative power absorbed.}
    \label{fig:cycle-mechanism}
\end{figure}

To see how the spine and the legs share work within the cycle, Figure~\ref{fig:cycle-mechanism} compares two trials with similar mean electrical power, impedance-controlled SPARC at 0.818~m/s and 58.75~W against the rigid spine at 0.673~m/s and 53.46~W.

Over each cycle, the spine supplies about 1.16~J of mechanical energy and absorbs about 2.17~J, so it takes in more than it returns. The lower subplot resolves this exchange. The spine absorbs energy as its axial velocity reverses, then supplies it as the fore and hind bodies separate, and the fore-body load and hind-body inertia reset it for the next cycle. This cycling reshapes the timing of the leg work.

The legs draw almost the same mean magnitude of mechanical power in the two trials, 22.76~W for impedance-controlled SPARC and 23.70~W for the rigid spine, but the upper subplot shows two different profiles. The rigid-spine trial concentrates leg power into sharp peaks separated by near-zero intervals, while the impedance-controlled SPARC trial spreads it more evenly through the cycle. The spine therefore redistributes the leg work within the cycle without increasing it.

\begin{figure}[t]
    \centering
    \includegraphics[width=\linewidth]{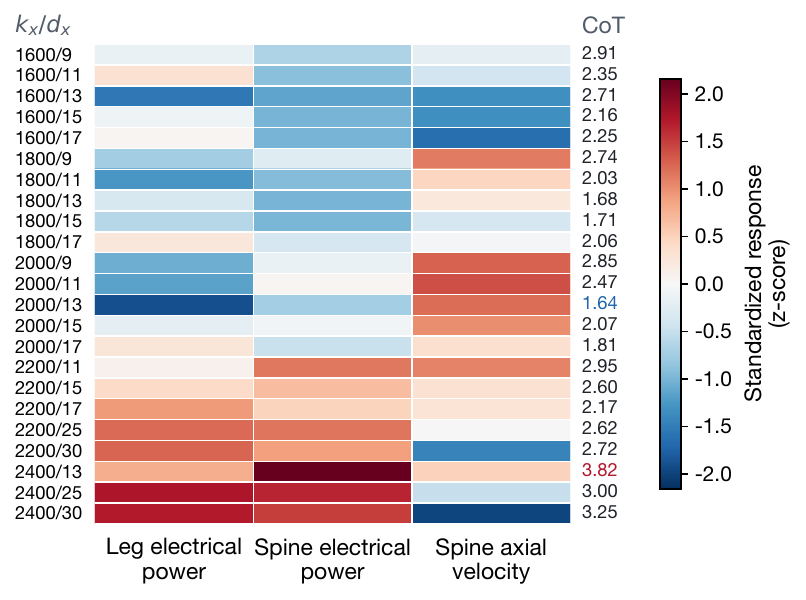}
    \caption{Measured response from 23 combinations of the spine parameters $k_x$ and $d_x$. The heatmap shows standardized leg electrical power, spine electrical power, and axial spine velocity, and the raw total electrical CoT is listed at right.}
    \label{fig:spine-fingerprint}
\end{figure}

\subsection{Impedance Parameter Sensitivity}\label{sec:sensitivity}
To isolate the effect of the axial impedance on locomotion, we swept the spine parameters over 23 settings and recorded one run at each. All 23 runs share the same gait and the same rotational impedance, with a 3.0~Hz bounding frequency, a 0.12~m stride for both leg pairs, and identical leg gains, so only $k_x$ and $d_x$ differ between them. Figure~\ref{fig:spine-fingerprint} shows that the commanded impedance changes both the axial motion and the split of electrical demand between the spine and the legs. At the stiff and heavily damped setting $k_x/d_x=2400/30$, both leg and spine electrical power are high while axial spine velocity is suppressed, giving a total electrical CoT of 3.25. The actuators spend substantial power holding the commanded impedance without producing axial motion that contributes to progression. Lowering the damping at the same stiffness does release the axial motion. At $2400/13$, spine electrical demand and axial velocity both increase, yet this setting draws the most power in the sweep (135~W) and is also the slowest (0.541~m/s), giving the highest CoT (3.82). The spine oscillation is poorly phased with the bounding cycle, so part of the actuator work does not add to forward motion.

The intermediate settings $2000/13$ and $1800/13$ produce the two lowest CoTs, 1.64 and 1.68, and the two highest speeds, 0.884 and 0.866~m/s. Here the spine and the legs work together, and both draw less electrical power while the robot moves faster. The rendered impedance therefore has to be matched to the gait dynamics, and a mismatched setting such as $2400/13$ draws 43\% more power while running 39\% slower.

\section{CONCLUSION}\label{sec:conclusion}

We presented SPARC, a compact spine module that combines active revolute and prismatic motion with task-space impedance control. Benchtop tests confirmed accurate rendering of the commanded axial stiffness and damping. We then evaluated bounding with impedance-controlled SPARC, position-controlled SPARC, and a 276~g rigid spine. Impedance-controlled SPARC reached 1.029~m/s, 1.34$\times$ the maximum speed of position-controlled SPARC and 1.53$\times$ that of the rigid spine. The faster runs were those in which the spine moved most and exchanged more mechanical power. Within a bounding cycle, the spine absorbed more mechanical energy than it returned, while its motion changed the timing of leg work.

The cost of higher speed is increased electrical energy consumption. Adding passive elasticity in parallel with the actuators could reduce the electrical demand while retaining active impedance regulation. Parallel stiffness has already been shown to reduce the power consumption of a bounding quadruped with a flexible spine~\cite{folkertsma2012parallel}. For SPARC, a passive element in parallel with the actuators could carry the baseline restoring force while the actuators regulate what remains. In addition, a limitation of this work is that we compared complete robot systems rather than isolating compliance, since each condition was tuned separately and the configuration with the rigid spine also differs in mass and actuator count. Future work will develop a controller to automatically tune the impedance parameters and extend the evaluation to a broader range of motions.

\balance
\bibliographystyle{IEEEtran}
\bibliography{references}






\end{document}